\documentclass[]{Ateam}

\usepackage{microtype}
\usepackage{graphicx}
\usepackage{subcaption}
\usepackage{afterpage}
\usepackage{tabularx}
\usepackage{booktabs}
\usepackage{multirow}
\usepackage{colortbl}
\usepackage{enumitem}
\usepackage{wrapfig}
\usepackage{nicefrac}
\usepackage{makecell}
\usepackage{xcolor}
\usepackage{pifont}
\usepackage{adjustbox}
\usepackage{amsmath}
\usepackage{amssymb}
\usepackage{mathtools}
\usepackage{amsthm}
\usepackage{xspace}

\newcommand{\maketitlesupplementary}{%
  \begin{center}
    \vspace{0.25em}
    \titlelist\par
  \end{center}
  \vspace{0.75em}
}

\title{$A_1$: A Fully Transparent Open-Source, Adaptive and Efficient Truncated Vision-Language-Action Model}

\author[1,3]{Kaidong Zhang*}
\author[2]{Jian Zhang*}
\author[2,3]{Rongtao Xu*$\dagger$}
\author[1]{Yu Sun}
\author[1]{Shuoshuo Xue}
\author[1]{Youpeng Wen}
\author[3]{Xiaoyu Guo}
\author[2]{Minghao Guo}
\author[1]{Weijia Liufu}
\author[3]{Liu Zihou}
\author[3]{Kangyi Ji}
\author[2]{Yangsong Zhang}
\author[3]{Jiarun Zhu}
\author[1]{Jingzhi Liu}
\author[1]{Zihang Li}
\author[1]{Ruiyi Chen}
\author[2]{Meng Cao}
\author[3]{Jingming Zhang}
\author[1]{Shen Zhao}
\author[2]{Xiaojun Chang}
\author[3]{Feng Zheng}
\author[2]{Ivan Laptev}
\author[1,2]{Xiaodan Liang$\ddagger$}

\affiliation[1]{\textbf{SYSU}}
\affiliation[2]{\textbf{MBZUAI}}
\affiliation[3]{\textbf{Spatialtemporal AI}}

\abstract{
Vision--Language--Action (VLA) models have emerged as a powerful paradigm for open-world robot manipulation, but their practical deployment is often constrained by \emph{cost}: billion-scale VLM backbones and iterative diffusion/flow-based action heads incur high latency and compute, making real-time control expensive on commodity hardware.
We present \textbf{$A_1$}, a \textbf{fully open-source and transparent} VLA framework designed for \emph{low-cost, high-throughput} inference without sacrificing manipulation success; Our approach leverages pretrained VLMs that provide implicit affordance priors for action generation. We release the full training stack (training code, data/data-processing pipeline, intermediate checkpoints, and evaluation scripts) to enable end-to-end reproducibility.
Beyond optimizing the VLM alone, $A_1$ targets the full inference pipeline by introducing a \textbf{budget-aware adaptive inference} scheme that jointly accelerates the \emph{backbone} and the \emph{action head}.
Specifically, we monitor \emph{action consistency} across intermediate VLM layers to trigger \textbf{early termination}, and propose \textbf{Inter-Layer Truncated Flow Matching} that \emph{warm-starts} denoising across layers, enabling accurate actions with substantially fewer effective denoising iterations.
Across simulation benchmarks (LIBERO, VLABench) and real robots (Franka, AgiBot), $A_1$ achieves state-of-the-art success rates while significantly reducing inference cost (e.g., up to \textbf{72\%} lower per-episode latency for flow-matching inference and up to \textbf{76.6\%} backbone computation reduction with minor performance degradation).
On RoboChallenge, $A_1$ achieves an average success rate of \textbf{29.00\%}, outperforming baselines including $\pi_0$ (28.33\%), X-VLA (21.33\%), and RDT-1B (15.00\%).

}

\contribution[*]{Equal contribution}
\contribution[\dagger]{Project Lead}
\contribution[\ddagger]{Correspondence}

\metadata[Code]{\url{https://github.com/ATeam-Research/A1}}
\metadata[Project Page]{\url{http://www.ateam.xin/\#/research/A1}}
\date{\today}
\begin{document}
\maketitle

\section{Introduction}
\label{sec:intro}
\begin{figure*}[t]
  \centering \includegraphics[width=1.0\linewidth]{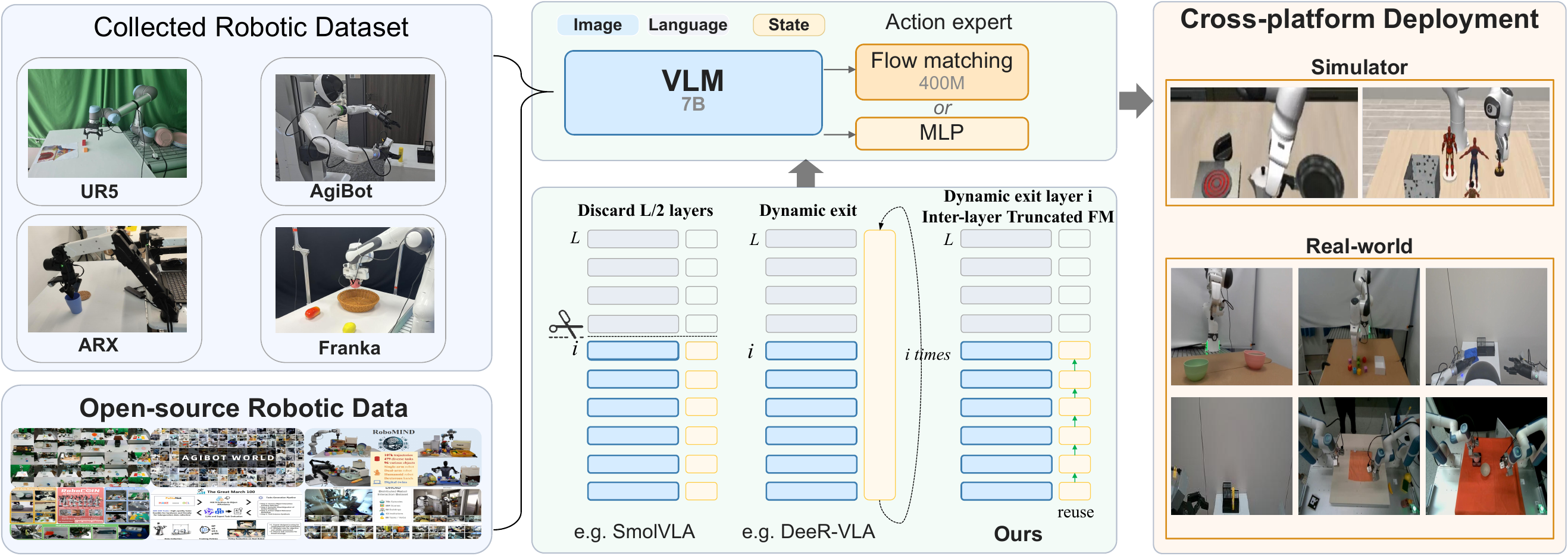}
  \caption{Overview of $A_1$. $A_1$ comprises a Vision Language Model (VLM) backbone and an action head, where the VLM provides semantically rich and affordance-aware representations for downstream action prediction. We instantiate the latter in two forms: a flow-matching head and an MLP head. To reduce end-to-end inference latency, we introduce a budget-aware acceleration scheme that is compatible with both action-head designs and jointly reduces backbone computation and action-head iterations. Extensive experiments in simulators, on real hardware, and on RoboChallenge demonstrate that $A_1$ achieves strong manipulation performance with substantially improved efficiency.}
  \label{fig:framework}
\end{figure*}

Robotic manipulation in the open world demands policies that can understand complex visual scenes and their underlying affordances~\cite{xu2025a0affordanceawarehierarchicalmodel}, 
and execute precise actions under tight latency budgets.
Vision--Language--Action (VLA) models have therefore become a dominant paradigm: a large-scale Vision--Language Model (VLM) compresses multimodal observations into a latent representation, and an action head---increasingly diffusion- or flow-matching--based---maps this latent into continuous motor commands.
This design inherits strong semantics from pretrained VLMs and expressive generative decoders, delivering impressive generalization across objects, instructions, and even robot morphologies.

However, this generality comes with a high \emph{deployment cost}.
State-of-the-art VLAs often rely on multi-billion-parameter backbones~\cite{black2024pi_0,intelligence2025pi05visionlanguageactionmodelopenworld,zhai2025ignitingvlmsembodiedspace}, while their diffusion/flow action heads typically require 10--20 iterative denoising steps.
Even if recent work reduces VLM latency via quantization~\cite{fang2025sqapvlasynergisticquantizationawarepruning}, sparsity~\cite{zhang2025molevladynamiclayerskippingvision}, or early-exit~\cite{yue2024deervladynamicinferencemultimodal}, the action head often remains untouched and quickly becomes the new bottleneck.
As a result, achieving \emph{real-time} control can require expensive hardware and substantial energy/compute budgets, limiting practical adoption.

In this paper, we introduce \textbf{$A_1$}, which is inspired by three empirical observations:
\begin{enumerate}
  \item \textbf{Trajectory convergence}: flow-matching trajectories can lock onto the correct mode within fewer than three denoising steps; additional iterations mostly refine precision with diminishing returns.
  \item \textbf{Action redundancy}: across consecutive control steps, many actions change smoothly and only require coarse updates \cite{zhang2024pivot}.
  \item \textbf{Layer-wise coupling}: intermediate VLM hidden states already encode sufficient spatial and visual features to seed the action prediction (e.g., the flow-matching vector field), making full-depth backbone evaluation often unnecessary.
\end{enumerate}

These observations point to a simple principle for \emph{low-cost, high-efficiency} VLA inference: \textbf{spend compute only when it changes the action}.
We therefore equip $A_1$ with a \textbf{budget-aware adaptive inference} mechanism.
At inference, we compute actions at intermediate VLM layers and perform an \textbf{action-consistency test} to decide whether to \textbf{terminate early}.
Crucially, to avoid shifting cost from the backbone to an iterative denoising head, we propose \textbf{Inter-Layer Truncated Flow Matching}:
we run only a small number of denoising steps per layer (e.g., $\delta{=}2$) and \emph{warm-start} the next layer's denoising from the previous layer’s prediction, rather than restarting from random noise.
This joint design accelerates \emph{both} components of the VLA pipeline, yielding substantial wall-clock savings (e.g., 37.8s$\rightarrow$10.5s per episode on LIBERO for flow-matching inference under our setup) while maintaining success rate (Table~\ref{tab:dynamic-fm}).

Beyond inference efficiency, we leverage pre-trained VLMs (Molmo~\cite{deitke2024molmopixmoopenweights}) that inherently capture affordance-aware representations for efficient action prediction~\cite{xu2025a0affordanceawarehierarchicalmodel}. $A_1$ is trained to generalize across robots and tasks using diverse robotic data. We pretrain $A_1$ using open-source robotic datasets including \textbf{DROID}~\cite{khazatsky2024droid}, \textbf{AgiBot}~\cite{agibotworldcontributors2025agibotworldcolosseolargescale}, \textbf{RoboCOIN}~\cite{wu2025robocoinopensourcedbimanualrobotic}, \textbf{RoboMind}~\cite{Wu_2025}, \textbf{GM-100}~\cite{wang2026greatmarch100100} and \textbf{RoboChallenge}~\cite{yakefu2025robochallengelargescalerealrobotevaluation}.
This practical training setup leverages publicly available data to support broad generalization without relying on proprietary large-scale corpora.

Extensive evaluations demonstrate that $A_1$ achieves strong manipulation performance across both simulation and real-world environments. On the RoboChallenge benchmark, $A_1$ surpasses open-source baselines including $\pi_0$ (28.33\%), X-VLA (21.33\%), and RDT-1B (15.00\%) with an average success rate of 29.00\%. In real-world experiments across four distinct robotic platforms (Franka, AgiBot, OpenArm, and Dobot-Arm), $A_1$ demonstrates strong performance with a mean success rate of 56.7\%, significantly outperforming baseline methods. In simulation, $A_1$ achieves competitive performance on LIBERO (96.6\%) and VLABench (53.5\%), demonstrating its robust generalization capabilities across diverse scenarios.

Furthermore, transparency and reproducibility are critical for sustained progress. We will open-source the model weights, training and inference code, data processing scripts/manifests, and evaluation protocols, so that the community can reproduce, audit, and extend our results.

Our contributions are:
\begin{itemize}
  \item \textbf{Joint acceleration of VLM backbone and action head:} a budget-aware adaptive inference scheme that simultaneously reduces redundant VLM computation via early-exit thresholding and cuts iterative action-head overhead via Inter-Layer Truncated Flow Matching with warm-start denoising, achieving substantial end-to-end latency reduction without performance degradation.
  \item \textbf{Scalable multi-robot pretraining:} pretraining on open-source robotic datasets plus 15{,}951 in-house trajectories across diverse robot platforms to support robust generalization.
  \item \textbf{Strong empirical results and fully open-source VLA:} achieving state-of-the-art manipulation performance including 29.00\% average success rate on RoboChallenge (outperforming $\pi_0$, X-VLA, and RDT-1B). We commit to releasing the full stack of artifacts for $A_1$ (model weights, training/inference code, data processing scripts/manifests, and evaluation protocols).
\end{itemize}
\section{Related Works}

\subsection{General Vision-Language-Action Frameworks}

Recent advancements in embodied AI have led to significant progress~\cite{ma2025phyblock,han2025multimodal,zhang2024navid,xu20253d,chen2025constraint,ren2024infiniteworld,liang2025structured}, particularly in the Vision-Language-Action (VLA) models. VLA models aim to unify perception, linguistic understanding, and control within a single multimodal policy, enabling general-purpose robotic reasoning and skill transfer. Some works adopt general VLA frameworks built upon Transformer or autoregressive architectures, which facilitate scalable pretraining and robust cross-task generalization~\cite{kimopenvla, goyal2023rvtroboticviewtransformer, goyal2024rvt2learningprecisemanipulation, chen2025internvlam1spatiallyguidedvisionlanguageaction, teamOctoOpenSourceGeneralist2024}. Beyond sequence modeling, recent studies introduce diffusion-based action models that leverage generative dynamics to produce temporally coherent, multimodal-conditioned policies~\cite{liuRDT1BDiffusionFoundation2024, reuss2024multimodaldiffusiontransformerlearning, dasari2024ingredientsroboticdiffusiontransformers, ze20243ddiffusionpolicygeneralizable}. These approaches extend traditional policy learning by formulating action generation as a stochastic denoising or prediction process, improving expressivity and stability in manipulation tasks. 

Complementary efforts focus on training and inference enhancement frameworks that augment VLA reasoning capabilities. Lightweight adapters~\cite{li2024generalistrobotpoliciesmatters} and trajectory-based prompting~\cite{zheng2025tracevlavisualtraceprompting} enhance spatial grounding and task adaptation, while embodied chain-of-thought reasoning~\cite{zawalski2025roboticcontrolembodiedchainofthought,zhang2025robridge,zhang2025mind,zhang2026robostereo} promotes interpretability through explicit action reasoning. Additionally, dual-system and verification-based designs~\cite{kwok2025robomonkeyscalingtesttimesampling, cui2025openhelixshortsurveyempirical} improve robustness and deployment reliability. Together, these developments mark the evolution of VLAs from general multimodal modeling toward efficient, interpretable, and embodied robotic intelligence.

\subsection{Efficient Vision-Language-Action Models}

With the rapid scaling of Vision-Language-Action (VLA) models, efficiency has become a central challenge for real-world deployment. EdgeVLA~\cite{budzianowski2025edgevlaefficientvisionlanguageactionmodels} accelerates inference by removing autoregressive dependencies and incorporating Small Language Models for edge deployment, while FAST~\cite{pertsch2025fastefficientactiontokenization} introduces frequency-space action tokenization for compact, high-frequency control. EfficientVLA~\cite{yang2025efficientvlatrainingfreeaccelerationcompression} and VLA-Cache~\cite{xu2025vlacacheefficientvisionlanguageactionmanipulation} further improve performance through training-free acceleration, structured layer pruning, and temporal token caching. Similarly, TinyVLA~\cite{wen2025tinyvlafastdataefficientvisionlanguageaction} and SmolVLA~\cite{shukor2025smolvlavisionlanguageactionmodelaffordable} design lightweight, data-efficient architectures for affordable and fast inference.  

Building on large-scale multimodal foundations, $\pi_0$~\cite{black2024pi_0} introduces a flow-matching architecture atop pre-trained VLMs for generalist robot control, while DeeR-VLA~\cite{yue2024deervladynamicinferencemultimodal} employs dynamic early-exit inference to adaptively scale computation under resource constraints. These models, together with EdgeVLA and EfficientVLA, exemplify the growing focus on balancing representational power with computational tractability. Our method differs from DeeR-VLA in that we employ a single shared action head during training, while at inference time we address the heavy computational cost caused by diffusion action heads across multiple time steps, leading to improved efficiency and performance. 



\section{Method}
\label{sec:method}


\begin{figure*}[t]
  \centering \includegraphics[width=0.98\textwidth]{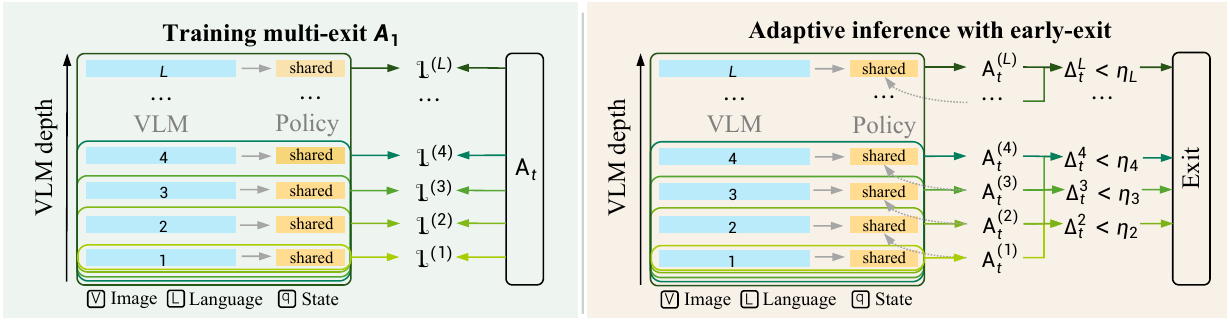} 
  \caption{Training and adaptive inference of $A_1$. During training, for each layer $i$ of VLM, the shared action head generates actions $A_t^{(i)}$. For flow matching, the model is executed with the same number of layers $i$ to generate predictions. All actions are supervised simultaneously. At inference time, we adaptively activate an appropriate size of VLM based on an exit criterion $c$. The threshold $\eta_i$ is generated by training set and $c$. When the VLM reaches the 
$i$-th layer, we computes the discrepancy between the current actions and those from the previous layer, and determines whether to terminate via $\eta_i$. We propose Inter-Layer Truncated Flow Matching to accelerate early-exit inference.} 
  \label{fig:post} 
\end{figure*}

\subsection{Overview}
As shown in Fig \ref{fig:framework}, Our $A_1$ architecture comprises a VLM and an action head. The VLM’s weights are initialized from Molmo \cite{deitke2024molmopixmoopenweights},  which endows the model with strong visual-semantic understanding as well as implicit affordance priors learned from large-scale multimodal pretraining. For the action head, we provide two implementations. One is Flow Matching \cite{lipman2023flowmatchinggenerativemodeling}, denoted as $A_1$-FM, which effectively represents high-dimensional action distributions. The other is an MLP head, referred to as $A_1$-MLP, which is supervised by L1 loss. It can quickly fit tasks and suppress noise \cite{kim2025finetuningvisionlanguageactionmodelsoptimizing}.

\subsection{Action Head Bridging for VLMs}
Formally, our goal is to learn the data distribution $p(\mathbf{A}_t|\mathbf{o}_t,\ell)$, where $\mathbf{o}_t=[\mathbf{I}_t^{1}, \ldots, \mathbf{I}_t^{n},\mathbf{q}_t]$ consists of the images from all of the cameras and the robot's proprioceptive state $\mathbf{q}_t$ (gripper pose, joint angles) at timestep $t$, $\ell$ is the language instruction. $\mathbf{A}_t=[\mathbf{a}_t,\mathbf{a}_{t+1},\ldots,\mathbf{a}_{t+H}] \in \mathbb{R}^{H\times D}$ is a predicted action chunk of future actions.

We implement two types of action modules that can be seamlessly integrated into VLMs: a flow-matching (FM) action expert \cite{black2024pi_0} and an MLP action head \cite{kim2025fine}.
For the flow-matching action expert, during training, we supervise these actions using a conditional flow matching loss \cite{lipman2023flow,black2024pi_0},
\begin{equation}
\mathcal{L}^{\tau}(\theta) = 
\mathbb{E}_{p(\mathbf{A}_t | \mathbf{o}_t),\, q(\mathbf{A}_t^{\tau} | \mathbf{A}_t)} 
\left\| 
\mathbf{v}_{\theta}(\mathbf{A}_t^{\tau}, \mathbf{o}_t) 
- \mathbf{u}(\mathbf{A}_t^{\tau} | \mathbf{A}_t) 
\right\|^2,
\end{equation}
where $\tau \in [0,1]$ denote flow matching timesteps. 
We sample random noise \(\boldsymbol{\epsilon} \sim \mathcal{N}(0, \mathbf{I})\), 
compute the “noisy actions” as \(\mathbf{A}_t^{\tau} = \tau \mathbf{A}_t + (1 - \tau)\boldsymbol{\epsilon}\), and then train the network outputs 
\(\mathbf{v}_{\theta}(\mathbf{A}_t^{\tau}, \mathbf{o}_t)\) to match the denoising vector field \(\mathbf{u}(\mathbf{A}_t^{\tau} | \mathbf{A}_t) = \boldsymbol{\epsilon} - \mathbf{A}_t\).
We follow $\pi_0$ for sampling the flow matching timestep $\tau$ from a beta distribution that emphasizes lower (noisier) timesteps.
At inference time, we generate actions by integrating the learned vector field from $\tau = 0$ to $\tau = 1$, starting with random noise $\mathbf{A}_{t}^{0} \sim \mathcal{N}(0, \mathbf{I})$. We use the forward Euler integration rule:
\begin{equation}
\mathbf{A}_{t}^{\tau+\delta} = \mathbf{A}_{t}^{\tau} + \delta \mathbf{v}_{\theta}(\mathbf{A}_{t}^{\tau}, \mathbf{o}_{t}),
\end{equation}
where $\delta$ is the integration step size.
We condition the action head via KV-conditioned self-attention: the prefix context produced by the main LLM is injected as past keys and values into a decoder-only stack, allowing the suffix tokens (action and state) to attend both to the cached prefix and to their own block. We use a Qwen3 model \cite{yang2025qwen3} with approximately 400M parameters as the FM action expert, and an additional MLP to output actions from the last hidden state.

For the MLP action head, Extra special query action tokens $S$ are added to input token IDs. Continuous actions $\hat{\mathbf{A}}_t$ are regressed from the hidden states $\mathbf{H}_t$ of extra dedicated action tokens with parallel decoding \cite{kim2025fine},
\begin{align}
    \mathbf{H}_t = f_{\phi}([\mathbf{o}_t,\ell, S])_{S}, \\ 
    \hat{\mathbf{A}}_t = g_{\psi}(\mathbf{H}_t). 
\end{align}
These actions are supervised by L1 loss,
\begin{equation}
    \mathcal{L}_{\text{MLP}}(\phi,\psi)
= \mathbb{E}_{p(\mathbf{A}_t \mid \mathbf{o}_t,\ell)}
\big\| \hat{\mathbf{A}}_t - \mathbf{A}_t \big\|.
\end{equation}

\subsection{Adaptive Inference Acceleration}
To use adaptive inference for acceleration, inspired by \cite{yue2024deer}, we introduce a simple and effective strategy during training stage. 
Specifically, we randomly sample a layer index $i \sim \mathcal{U}(0, L)$, 
where $L$ denotes the total number of layers in the main LLM. 
For the MLP action head, instead of using the last hidden state to generate actions as in the conventional design,  we supervise the loss $\mathcal{L}^{(i)}$ on the actions predicted from the hidden state at layer $i$. 
For the FM-based action head, the main LLM is executed up to layer $i$, 
and the FM action head is executed for the corresponding number of layers $i$, 
with the loss computed accordingly. In other words, the LLM does not need to run through all layers; it can be bridged to the action head after executing up to layer $i$. To achieve more stable training, another approach is to supervise the action loss from all $L$ layers.
\subsubsection{Early-Termination Inference via Action-Consistency Thresholding}
We use the early-exit mechanism inspired by \cite{yue2024deer} for large VLA inference that adaptively terminates the forward pass when the predicted action chunk stabilizes across layers. Early exit at layer $i$ is triggered by a consistency test against the previous layer’s action:
\begin{equation}
 \Delta_t^{i}=d(\mathbf{A}_t^{(i)}, \mathbf{A}_t^{(i-1)}) < \eta_i
 \label{eq:delta}
\end{equation}
where d(·,·) is a discrepancy metric and $\eta_i$ is a layer-specific threshold calibrated offline. superscript $(i)$ denotes action chunk generated from layer $i$.
We consider several vector metrics to measure action stability between successive exits, including cosine similarity, L2 distance and mean absolute deviation. 

The threshold $\eta_i$ is calibrated from the training data and determined by the probability distribution.
Given the training set $\mathcal{D}$, we run a single forward pass and collect layerwise action discrepancies. 
For each eligible exit layer $i$ in an ordered set $\mathcal{E} = \{ i_1 < i_2 < \ldots < i_K \}$, we compute: $S_i = \{ \Delta_t^{i} \mid (x_t, \ldots) \in \mathcal{D} \}.$ Stacking across exits yields a matrix of empirical discrepancies:
\begin{equation}
    V \in \mathbb{R}^{K \times N}, \quad V[k, n] \equiv \Delta_{t_n}^{i_k},
\end{equation}
where $K\leq L$ and $N$ is the total number of samples collected over $\mathcal{D}$. 
This ``values'' matrix compactly captures how much actions change from 
layer $i_{k-1}$ to $i_k$ across the dataset.

We translate a desired compute budget into a target early-exit probability mass over exits, 
$\mathbf{p} = (p_1, \ldots, p_K)$, using a parametric family \textit{exit\_dist} with a exit criterion $c$. 
We adopt an exponential distribution that emphasize earlier exits for stronger savings,
\begin{equation}
        p_k \propto \rho^k, \quad \rho = c > 0.
\label{eq:exit_exp}
\end{equation}
We normalize $\mathbf{p}$ to sum to 1. Smaller \textit{exit\_ratio} in the exponential family favors earlier exits. Gaussian and Gamma distribution provide symmetric and skewed allocations, respectively. More details can be found in the supplementary material.

Given $\mathbf{p}$, we set thresholds $\{\eta_{i_k}\}$ by selecting per-layer quantiles of discrepancies 
so that approximately a fraction $p_k$ of the ``remaining'' samples will exit at layer $i_k$. 
Concretely, we proceed from early to late exits and at each step choose $\eta_{i_k}$ as the 
$p_k$-quantile of the unassigned portion of $V[k, :]$. The last exit uses $\eta_{i_{K}} = +\infty$ (always exit), 
ensuring a proper fallback.

Formally, let $\mathcal{I}$ be the index set of samples not yet assigned to any earlier exit. We set
\begin{equation}
    \eta_{i_k} = Q_{p_k}\!\big( \{ V[k, n] \}_{n \in \mathcal{I}} \big), 
\quad 
\mathcal{I} \leftarrow \{ n \in \mathcal{I} \mid V[k, n] > \eta_{i_k} \},
\label{eq:eta_i}
\end{equation}
where $Q$ denotes the quantile operator. $Q_{p_k}\!\big( \{ V[k, n] \}_{n \in \mathcal{I}} \big)$ returns the $p_k$-quantile of the discrepancy values at exit $k$ over the unassigned sample index set $\mathcal{I}$.

This filtered-quantile procedure enforces disjoint assignment across exits 
while matching the target exit proportions.

    
    

\subsubsection{Inter-Layer Truncated Flow Matching}
For normal action inference, the main LLM and the action head are executed both once. However, for early-exit inference, although the main LLM runs fewer layers, the action head needs to be executed at every layer until early exit, which increases the computational cost. This is acceptable for an MLP-based action head, since its computation is negligible compared to the LLM. However, for FM-based action head, a $\delta$-step denoising process (typically $\delta$ = 10 or 20) is required, which adds overhead and inference time. 

To address this issue, we propose Inter-Layer Truncated Flow Matching. We first set the number of denoising steps $\delta$ to a small value (e.g., 2). The LLM is executed layer by layer. At each layer $i$, the FM action head performs $\delta$ denoising steps to generate an action chunk $\mathbf{A}_{t}^{(i)}$, which is then compared with the $\mathbf{A}_{t}^{(i-1)}$ from the previous layer to determine whether to exit early via Eq.~\eqref{eq:delta}. The output of the current layer is passed to the next layer as the initial condition for denoising, as shown in
\begin{equation}
    \mathbf{A}_{t}^{0(i+1)} = \mathbf{A}_{t}^{1(i)},
    \label{eq:anchor}
\end{equation}
instead of starting from random noise. This allows the denoising process to be propagated across layers, or in other words, provides a \textbf{warm-start initialization} for denoising models at different layer depths. This strategy significantly reduces computational cost while maintaining accuracy.


\section{Training Recipe}
We adopt a two-stage training pipeline.
In the first stage, we pre-train the Vision--Language--Action (VLA) model using large-scale robotic datasets.
In the second stage, we fine-tune the model for specific robot embodiments or downstream tasks.
We detail the data composition and training procedure below.






\subsection{Pre-Training Data Composition}
\label{sec:pretrain-data}
We build a large-scale \emph{robot-trajectory-only} pre-training corpus for $A_1$, consisting exclusively of robotic demonstrations from diverse embodiments and environments, focusing on practical multi-robot generalization.

\textbf{Open-source Robotic Dataset.}
We leverage publicly available robotic datasets including \textbf{DROID}~\cite{khazatsky2024droid}, \textbf{AgiBot}~\cite{agibotworldcontributors2025agibotworldcolosseolargescale}, \textbf{RoboCOIN}~\cite{wu2025robocoinopensourcedbimanualrobotic}, \textbf{RoboMind}~\cite{Wu_2025}, \textbf{GM-100}~\cite{wang2026greatmarch100100} and \textbf{RoboChallenge}~\cite{yakefu2025robochallengelargescalerealrobotevaluation}.
These sources provide heterogeneous robot morphologies (e.g., different manipulators and sensor setups), diverse task families (e.g., tabletop manipulation and articulated-object interaction), and varied scene statistics, which together encourage broad generalization.

\textbf{Collected Robotic Dataset.}
To enable effective deployment on our target platforms, we collect \textbf{15{,}951} real-world trajectories on multiple robots, including \textbf{ARX}, \textbf{Franka}, \textbf{UR5}, and \textbf{Agibot}.
Models pre-trained solely on open-source datasets often exhibit significant performance degradation when directly deployed in our local setups, due to differences in hardware configurations, control interfaces, sensing pipelines, and environment distributions.
The collected dataset reflects our target deployment conditions, including consistent sensor setups, control frequencies, and action parameterizations.
It thus serves as a deployment-aligned data source that adapts the pre-training distribution toward our local domain, reducing the cross-platform gap.

\textbf{Unified representation and quality control.}
All datasets are converted into a consistent episodic format:
each trajectory is represented as a sequence of synchronized tuples
$(o_t, s_t, \ell, a_t)$,
where $o_t$ denotes visual observations (e.g., RGB images, potentially multi-view when available), $s_t$ denotes robot state/proprioception when available, $\ell$ is a language goal, and $a_t$ is the continuous action.
We perform lightweight filtering to remove corrupted episodes (missing frames/timestamps), extreme outliers, and overly redundant segments (e.g., long idle prefixes), and apply balanced sampling across sources to prevent a single dataset or robot from dominating training.

\subsection{Training Procedure}

\textbf{Training Pipeline.} 
Our training procedure consists of two stages: large-scale pre-training on diverse real-world robot datasets, followed by task-specific fine-tuning. In the pre-training stage, the model is trained end-to-end on a mixture of self-collected teleoperation data and open-source embodiment datasets to acquire generalizable manipulation priors. Subsequently, we fine-tune the pre-trained checkpoint on downstream tasks with smaller, high-quality demonstration datasets to adapt the policy to specific skills and environmental constraints.

\textbf{Data Processing and Augmentation.} 
We apply aggressive data augmentation strategies to improve robustness and generalization. For visual inputs, we employ image sharpening and random erasing to enhance texture details and prevent overfitting to background distractors. For proprioceptive states, we utilize action augmentation by randomly masking out state dimensions (state zero-out) to improve robustness against partial observability. Notably, we intentionally avoid state normalization across datasets to preserve the intrinsic action space characteristics of identical robot embodiments, ensuring that the model learns consistent physical dynamics rather than normalized abstractions. Additionally, we filter out static frames and low-velocity segments to eliminate redundant timesteps and focus the learning on meaningful motion primitives.

\textbf{Data Sampling Strategy.} 
To ensure balanced learning across heterogeneous data sources, we implement a hierarchical sampling strategy. First, we apply dataset-level balanced sampling, where each dataset is sampled with equal probability to prevent the model from overfitting to any single data distribution. Within each dataset, we further enforce embodiment-balanced sampling, ensuring that different robot morphologies contribute equally to each training batch. This two-level balancing mechanism guarantees exposure to diverse hardware configurations and task distributions while mitigating bias toward over-represented robots or environments.

\textbf{Optimization and Learning Rate Scheduling.} 
During training, we freeze the Vision Transformer (ViT) backbone to preserve the pre-trained visual representations. The Vision-Language Model (VLM) components are optimized with a learning rate of $5\times10^{-5}$, while the action head employs a higher learning rate of $5\times10^{-4}$ to facilitate rapid adaptation to motor control objectives. We employ a warm-up strategy during the initial training steps, linearly increasing the learning rate from zero to the target value while keeping the VLM backbone frozen (zero learning rate) during the first 1,000 steps. After warm-up, we apply cosine annealing to gradually decay the learning rate, ensuring stable convergence and preventing catastrophic forgetting of pre-trained capabilities.

\begin{table*}[t]
    \centering
    \small
    \caption{Experimental Results of Simulation Benchmarks (\%). }
    \begin{tabular}{l|ccccc|ccccc}
    \toprule
        Model &  \multicolumn{5}{c|}{LIBERO} & \multicolumn{5}{c}{VLABench}\\ 
        & Spatial & Object & Goal & Long & Avg. & Toy & Fruit & Painting & Mahjong & Avg.\\
         \midrule 
         Octo \cite{teamOctoOpenSourceGeneralist2024} & 78.9  & 85.7 & 84.6 & 51.1 & 75.1 & 0 & 0 & 6 &  0 & 1.5 \\
            OpenVLA \cite{kimopenvla} & 84.7 & 88.4 & 79.2 & 53.7 & 76.5 &  4 & 6 & 40 &  8 & 14.5 \\
            OpenVLA-OFT \cite{kim2025finetuningvisionlanguageactionmodelsoptimizing}& 97.6 & 98.4 & 97.9 & $\mathbf{94.5}$ & $\mathbf{97.1}$ & -  & - & - & -  & - \\
            CoT-VLA \cite{zhao2025cotvlavisualchainofthoughtreasoning}& 87.5 & 91.6 & 87.6 & 69.0 & 81.1 & - & - & - &  - & - \\
            MolmoAct \cite{lee2025molmoact}& 87.0 & 95.4 & 87.6 & 77.2 & 86.6 & - & - & - & -  & - \\
            SmolVLA \cite{shukor2025smolvlavisionlanguageactionmodelaffordable}& 93.0 & 94.0 & 91.0 & 77.0 & 88.8 & - & - & - &  - & - \\
        $\pi_0$ \cite{black2024pi_0}& 96.8 & 98.8 & 95.8 & 85.2 & 94.2 & 52 & 60 & 24  & $\mathbf{32}$  & 42 \\
        $\pi_{0.5}$ \cite{intelligence2025pi05visionlanguageactionmodelopenworld}& $\mathbf{98.8}$ & 98.2 & $\mathbf{98.0}$ & 92.4 & 96.9 & $\mathbf{70}$ & 62 & 44 & 22 & 49.5 \\ \midrule 
        $A_1$ & 97.4 &  $\mathbf{99.8}$ & 97.6 & 91.4 & 96.6 & 62 & $\mathbf{64}$ & $\mathbf{70}$ & 18 & $\mathbf{53.5}$ \\
    \bottomrule
    \end{tabular}
    
    \vspace{-0.5em}
    \label{tab:simulation}
\end{table*}

\section{Experiments}
We first introduce the experimental setup \ref{exp:experimental_setup} along with the
benchmarks, baselines and hardware settings. We present the main results
in real world \ref{exp:real_world} and simulation \ref{exp:simulation}. Additionally, we conduct ablation studies in section \ref{exp:ablation}.

\begin{figure*}[!ht]
\begin{center}
\includegraphics[width=1\linewidth]{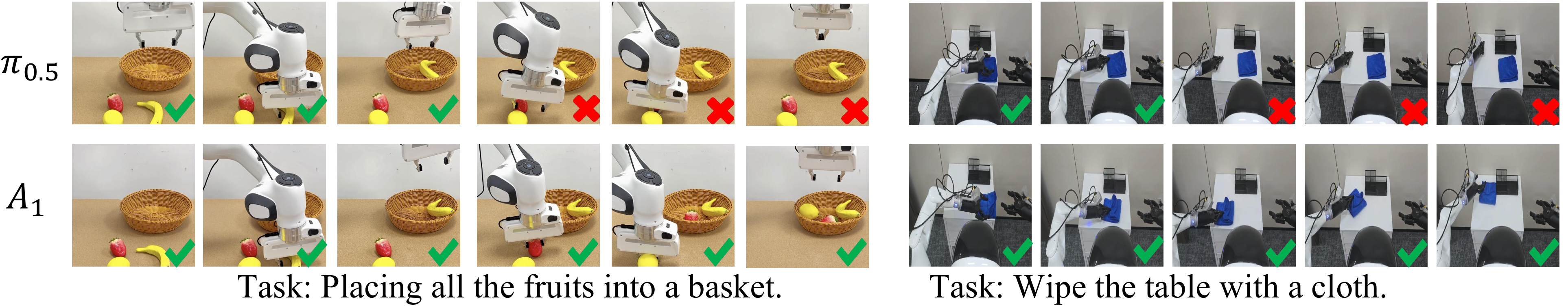}

\end{center}
   \caption{ Demonstrations show the execution process of $A_1$ (second row) and baselines $\pi_{0.5}$ (first row).
   }
\label{fig:real_demo}
\end{figure*}


\begin{table*}[ht]
\centering
\footnotesize
\setlength{\tabcolsep}{3pt}
\caption{Experimental Results of Real-World Evaluation (\%) across Multiple Robot Platforms.}
\label{tab:real_world}
\resizebox{\textwidth}{!}{
\begin{tabular}{l|cc|cccc|cc|cc|cc|c}
\toprule
\multirow{2}{*}{Model} & \multicolumn{2}{c|}{UR5} & \multicolumn{4}{c|}{Franka} & \multicolumn{2}{c|}{AgiBot} & \multicolumn{2}{c|}{OpenArm} & \multicolumn{2}{c|}{Dobot-Arm} & \multirow{2}{*}{Mean} \\
\cmidrule(lr){2-3}\cmidrule(lr){4-7}\cmidrule(lr){8-9}\cmidrule(lr){10-11}\cmidrule(lr){12-13}
& stack & arrange & put cup & arrange & fruits & move & pick & clean & tidy & select & cook & pour & \\
& blocks & fruits & on coaster & fruits & (small) & objects & glue & table & up & yellow & vegetable & water & \\
\midrule
$\pi_0$  \cite{black2024pi_0}   & $\mathbf{100}$ & 80 & $\mathbf{70}$ & 30 & 10 & 10 & 30 & 0 & 30 & $\mathbf{80}$ & 10 & $\mathbf{40}$ & 40.8\\
$\pi_{0.5}$ \cite{intelligence2025pi05visionlanguageactionmodelopenworld}& 80 & $\mathbf{100}$ & 50 & 20 & 30 & 40 & 60 & 10 & $\mathbf{80}$ & 60 & $\mathbf{20}$ & 20 & 47.5\\
$A_1$       & $\mathbf{100}$ & 60 & 50 & $\mathbf{40}$ & $\mathbf{50}$ & $\mathbf{80}$ & $\mathbf{80}$ & $\mathbf{20}$ & $\mathbf{80}$ & 70 & $\mathbf{20}$ & 30 & $\mathbf{56.7}$\\

\bottomrule
\end{tabular}
}
\end{table*}

\subsection{Experiment Settings}
\label{exp:experimental_setup}
\textbf{Simulation Benchmasks.} We conducted experiments in
two simulation environments: LIBERO \cite{liu2023libero} and VLABench \cite{zhang2025vlabench}. LIBERO is a robotics manipulation simulation benchmark for lifelong learning, consisting of four task suites: Spatial, Object, Goal, and Long. LIBERO-Long extends the manipulation chain to 5–10 steps.
VLABench is an open "language-condition-operation" benchmark for large models, emphasizing the deep integration of world knowledge, common sense, and multi-step reasoning. We selected four tasks on VLABench that fully examine the model's visual language understanding capabilities.
For LIBERO, following  \cite{kim2025finetuningvisionlanguageactionmodelsoptimizing}, we report the average success rate over 500 trials per task suite; for VLABench, we report the mean success rate over 50 trials per task.

\noindent\textbf{Real-world Robots and Tasks Setup.}  
We conduct extensive real-world evaluations across four distinct robotic platforms: Franka, AgiBot, OpenArm, and Dobot-Arm. Our evaluation suite comprises seven diverse manipulation tasks: (1) placing a cup on a white coaster, (2) arranging fruits into a basket, (3) stacking color blocks, (4) picking and storing glue, (5) wiping the table with a cloth, (6) tidying up objects, and (7) cooking vegetables. For few-shot learning evaluation, we specifically collected a small dataset containing only 50 samples for the fruit arrangement task. In total, over 3,000 trajectories were collected across all platforms, with each task tested 10 times. To further assess generalization capabilities, we additionally evaluate on the RoboChallenge \cite{yakefu2025robochallengelargescalerealrobotevaluation} benchmark comprising 30 real-robot tasks spanning multiple embodiments.


\subsection{Model Computational Cost Analysis}
Molmo-7B VLM consists of a vision encoder (e.g., CLIP, SigLIP) and a 28-layer Qwen2-7B. The total inference cost is 11,074.39 GFLOPs (sequence length 352), with CLIP accounting for 2,013.36 GFLOPs and each LLM layer for 323.61 GFLOPs. With an action dimension of 7 and a chunk size of 8, the MLP action head requires 1.850 GFLOPs. Flow matching with Qwen3-400M costs 0.493 GFLOPs per timestep (4.931 GFLOPs for 10 steps). $A_1$-FM normal inference requires 11.130 TFLOPs. 
For early-exit inference, when $A_1$-FM runs to the final layer, each layer requires $\delta$ = 10 denoising steps. Including computation and threshold comparison, this results in 11.160 TFLOPs and a 4.44 s inference time. In contrast, when $\delta$ = 2, the inference time is only 0.73 s. This indicates that although the flow-matching model has a small computation cost, the iterative denoising steps at each layer incur a substantial computational time.
\begin{table}[h]
\centering
\caption{Computation cost and latency of different parts of the model. $A_1$-FM$^{e}$ denotes adaptive early-exit reaching the final layer, where we set $K$=14 to evaluate the exit criterion (Eq.~(6)) every two layers.}
\label{tab:time_analysis}
\begin{tabular}{l|c|c|c|c|c|c}
\toprule
  & CLIP & \makecell{LLM \\ ($L$=28)} & \makecell{FM \\ ($\delta$=10)} &
  \makecell{$A_1$-FM \\ ($\delta$=10)} &
  \makecell{$A_1$-FM$^{e}$ \\ ($\delta$=10)} &
  \makecell{$A_1$-FM$^{e}$ \\ ($\delta$=2)} \\
\midrule
Time (s) & 0.167 & 0.612 & 0.366 & 1.151 & 4.443 & 0.728 \\
GFLOPS   & 2013.36 & 9061.01 & 4.93 & 11130.30 & 11503.30 & 11160.20 \\
\bottomrule
\end{tabular}
\end{table}

\subsection{Simulation Benchmark Results}
\label{exp:simulation}

As shown in Table \ref{tab:simulation}, our method achieves a success rate of 96.6\% on LIBERO and 99.8\% on the OBJECT task. On VLABench, $A_1$ achieves an average success rate of 53.5\%, 4\% higher than $\pi_{0.5}$. $A_1$ accurately identifies the task target on VLABench tasks, with most failures due to objects falling. We found that even when the robotic arm's gripping deviation causes significant changes in the object's pose, the model can still complete the task, indicating that our model is able to recognize the task objective rather than simply fitting a trajectory.

\subsection{Real-World Experiment Results}
\label{exp:real_world}

\textbf{Real-world Manipulation.} As shown in Table~\ref{tab:real_world}, our model $A_1$ achieves a \textbf{56.7\%} average success rate, outperforming $\pi_{0.5}$ (47.5\%) and $\pi_0$ (40.8\%) by \textbf{9.2\%} and \textbf{15.9\%}, respectively.

$A_1$ demonstrates superior performance in both fine manipulation and long-horizon tasks. For instance, on AgiBot's ``pick glue'' task, $A_1$ attains \textbf{80\%} success (vs. 60\% and 30\%), while on ``clean table'' it reaches \textbf{20\%} (vs. 10\% and 0\%). Notably, with only 50 samples on ``fruits (small)'', $A_1$ achieves \textbf{50\%} success, surpassing baselines by 20--40\%.

Figure~\ref{fig:real_demo} presents qualitative results: $\pi_{0.5}$ often grasps between objects or closes the gripper prematurely, whereas $A_1$ executes actions more accurately without being distracted by multiple objects.

\begin{figure*}[h]
\begin{center}
\includegraphics[width=1\linewidth]{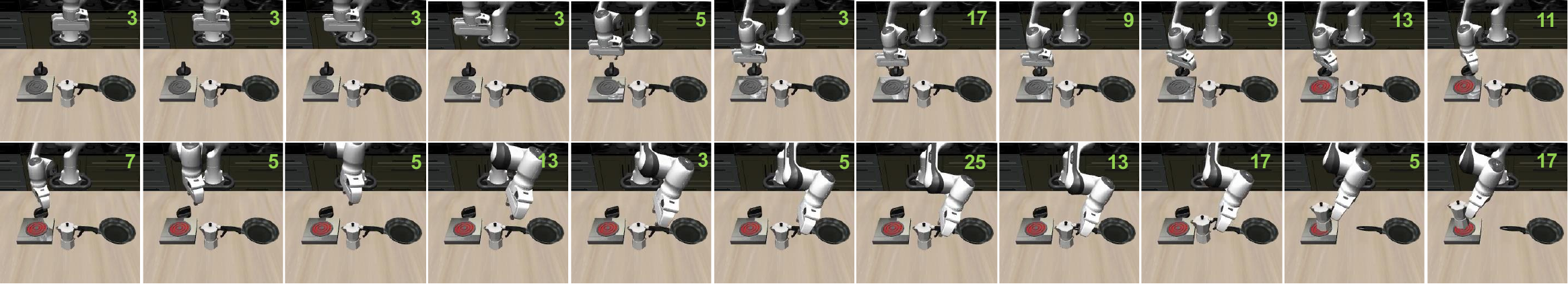}
\vspace{-3mm}
\end{center}
   \caption{Example of adaptive inference visualization for the $A_1$ model with the exit criterion $c=0.6$. A successful execution episode from the LIBERO-Long task (instruction: \textit{turn on the stove and put the moka pot on it}). Green numbers indicate the layer indices where the $A_1$ model exits (the model has 28 layers in total). At each frame, the model outputs actions for the next 8 time steps (action chunk = 8). 
   }
\label{fig:vis_exit}
\end{figure*}

\textbf{RoboChallenge Results.}
On the RoboChallenge Table30 benchmark, we present the results of $A_1$ in Table~\ref{tab:same_size_bmk_sr_transposed}. While state-of-the-art VLA systems often rely on closed-source data or proprietary training pipelines, A1 breaks this paradigm: as a completely transparent system with no dependencies on closed-source components, it achieves an average success rate of \textbf{29.00\%, ranking sixth} overall. This performance significantly surpasses comparable open-source baselines including $\pi_0$ (28.33\%), X-VLA (21.33\%), and RDT-1B (15.00\%), demonstrating that transparent and reproducible open research is highly competitive in real-world robotic manipulation tasks.

Specifically, A1 achieves high success rates on critical precise manipulation and long-horizon manipulation tasks such as \textbf{``Open Drawer'' (100\%), ``Put Cup on Coaster'' (90\%), and ``Stack Bowls'' (80\%)}, these results demonstrate the practical viability of open-source solutions =in achieving reliable task execution on complex real-robot challenges, providing a fully reproducible technical pathway for low-cost, high-transparency robotic policy deployment.

\subsection{Ablation Study}
\label{exp:ablation}
\subsubsection{Effectiveness of Early-Termination Inference}
As shown in Table \ref{tab:dynamic-exit}, we set different value of exit criterion $c$ based on exponential distribution (Eqs.~(\ref{eq:exit_exp}) and (\ref{eq:eta_i})) for adaptive inference. 
When $c$ = 1.0 , the $A_1$-MLP model achieves its best performance, with an average success rate of 96.6\%, while reducing layer computation by 15.6\% compared to full inference. As $c$ decreases to 0.7, 0.4, and 0.1, the computational reduction increases to 39.1\%, 58.5\%, and 76.6\%, respectively, while the success rate drops by only 0.3\%, 2.3\% and 1.7\%. Even when reducing 76.6\% of the computation cost, the model still achieves a 92.3\% success rate. This indicates that a large portion of computation in VLM layers is redundant, and dynamic inference enables the model to adaptively select effective features. Therefore the accuracy of multi-exit training and adaptive inference with $c$=1.0 is better than full-layer training and inference. 

Interestingly, less computation can sometimes lead to better performance. For example, on the LIBERO-\textit{Spatial} task, with $c$ = 0.7, the model achieves the highest success rate of 98.4\%.
These results demonstrate that the model can adaptively select the most effective features. The visualization analysis in Figure \ref{fig:vis_exit} also supports this observation. For most simple actions such as movement, the model exits early (e.g., at layer 3 or 5). For key complex actions such as turning on the stove or picking up the pot, the model proceeds to deeper layers (e.g., 17 or 25) to produce more accurate actions.

Table \ref{tab:dynamic-fm} demonstrates the effectiveness of early-exit inference on flow-matching–based models, significantly reducing computation while maintaining accuracy. 

\begin{table*}[htbp]
  \centering
  \scriptsize
  \caption{Comparison with state-of-the-art open-source VLA models on Table30 benchmark in RoboChallenge. List in success rate. \textbf{Bold} for highest. \textsuperscript{§}Denotes fully open-source models with complete training stack, weights, and data pipelines.}
  \label{tab:same_size_bmk_sr_transposed}
  
  \begin{tabularx}{\textwidth}{@{}>{\raggedright\arraybackslash}p{2cm}|*{10}{>{\centering\arraybackslash}X}@{}}
    \toprule
    \textbf{Model} & \textbf{Arrange} & \textbf{Hang} & \textbf{Set} & \textbf{Shred} & \textbf{Sort} & \textbf{Stack} & \textbf{Move} & \textbf{Press} & \textbf{Arrange} & \textbf{Arrange} \\
    & \textbf{Fruits} & \textbf{Cup} & \textbf{Plates} & \textbf{Paper} & \textbf{Books} & \textbf{Blocks} & \textbf{Objects} & \textbf{Buttons} & \textbf{Flowers} & \textbf{Cups} \\
    \midrule
    DM0 \cite{yu2026dm0embodiednativevisionlanguageactionmodel}& \textbf{100} & \textbf{80} & \textbf{100} & \textbf{30} & \textbf{20} & \textbf{100} & \textbf{100} & 90 & \textbf{70} & 30 \\
    Spirit-v1.5 \cite{spiritai2026spiritv15}& 80 & \textbf{80} & 80 & 20 & 0 & 80 & 80 & 90 & 50 & 0 \\
    GigaBrain \cite{gigabrainteam2025gigabrain0worldmodelpoweredvisionlanguageaction}& 60 & 40 & 90 & 0 & 0 & \textbf{100} & 90 & 40 & 40 & \textbf{80} \\
    $\pi_{0.5}$ \cite{intelligence2025pi05visionlanguageactionmodelopenworld}& 40 & 50 & 80 & 0 & 0 & \textbf{100} & 50 & 0 & 50 & 0 \\	
    wall-oss \cite{zhai2025ignitingvlmsembodiedspace}& 80 & 60 & 50 & 0 & 0 & \textbf{100} & 60 & \textbf{100} & 20 & 0 \\
    $A_1$\textsuperscript{§} & 60 & 60 & 30 & 20 & 0 & 60 & 30 & 0 & 10 & 20 \\
    $\pi_{0}$ \cite{black2024pi_0}& 20 & 50 & 10 & 30 & 0 & 70 & 50 & 0 & 50 & 0 \\
    X-VLA\textsuperscript{§} \cite{zheng2025xvlasoftpromptedtransformerscalable}& 0 & 0 & 0 & 0 & 0 & 0 & 20 & 90 & 0 & 0 \\
    RDT-1B\textsuperscript{§} \cite{liu2025rdt1bdiffusionfoundationmodel}& 0 & 0 & 0 & 0 & 0 & 10 & 50 & 0 & 10 & 10 \\
    \bottomrule
  \end{tabularx}
  
  \vspace{0.5em}
  
  \begin{tabularx}{\textwidth}{@{}>{\raggedright\arraybackslash}p{2cm}|*{10}{>{\centering\arraybackslash}X}@{}}
    \toprule
    \textbf{Model} & \textbf{Fold} & \textbf{Open} & \textbf{Place} & \textbf{Put Cup} & \textbf{Search} & \textbf{Sort} & \textbf{Turn} & \textbf{Water} & \textbf{Wipe} & \textbf{Clean} \\
    & \textbf{Cloth} & \textbf{Drawer} & \textbf{Shoes} & \textbf{ Coaster} & \textbf{Boxes} & \textbf{Elec.} & \textbf{Light} & \textbf{Plant} & \textbf{Table} & \textbf{Table} \\
    \midrule
    DM0 \cite{yu2026dm0embodiednativevisionlanguageactionmodel}& 20 & \textbf{100} & \textbf{100} & \textbf{100} & \textbf{100} & 0 & \textbf{80} & \textbf{80} & 0 & 0 \\
    Spirit-v1.5 \cite{spiritai2026spiritv15}& 20 & 70 & 90 & 90 & 90 & 30 & \textbf{80} & 0 & 0 & 30 \\
    GigaBrain \cite{gigabrainteam2025gigabrain0worldmodelpoweredvisionlanguageaction}& 10 & \textbf{100} & 50 & \textbf{100} & 80 & 0 & 60 & 60 & 0 & \textbf{40} \\
    $\pi_{0.5}$ & 20 & 40 & 90 & 90 & 80 & \textbf{50} & 40 & 0 & 0 & 10 \\
    wall-oss \cite{zhai2025ignitingvlmsembodiedspace}& 10 & 70 & 60 & 70 & 50 & 0 & 40 & 0 & 0 & 10 \\
    $A_1$\textsuperscript{§} & 10 & \textbf{100} & 60 & 90 & 50 & 0 & 50 & 0 & 0 & 0 \\
    $\pi_{0}$ \cite{black2024pi_0}& 0 & 0 & 80 & 60 & 70 & 0 & 10 & 0 & 0 & 0 \\
    X-VLA\textsuperscript{§} \cite{zheng2025xvlasoftpromptedtransformerscalable}& 10 & 0 & 50 & \textbf{100} & 30 & 20 & 0 & 0 & 0 & 0 \\
    RDT-1B\textsuperscript{§} \cite{liu2025rdt1bdiffusionfoundationmodel}& \textbf{30} & 70 & 60 & 80 & 10 & 0 & 20 & 0 & 0 & 0 \\
    \bottomrule
  \end{tabularx}
  
  \vspace{0.5em}
  
  \begin{tabularx}{\textwidth}{@{}>{\raggedright\arraybackslash}p{2cm}|*{10}{>{\centering\arraybackslash}X}|>{\centering\arraybackslash}X@{}}
    \toprule
    \textbf{Model} & \textbf{Make} & \textbf{Plug} & \textbf{Pour} & \textbf{Put} & \textbf{Put} & \textbf{Scan} & \textbf{Stack} & \textbf{Stick} & \textbf{Sweep} & \textbf{Turn} & \textbf{Mean} \\
    & \textbf{Sand.} & \textbf{Cable} & \textbf{Fries} & \textbf{Opener} & \textbf{Pen} & \textbf{QR} & \textbf{Bowls} & \textbf{Tape} & \textbf{Rub.} & \textbf{Faucet} & \\
    \midrule
    DM0 \cite{yu2026dm0embodiednativevisionlanguageactionmodel}& 0 & \textbf{80} & 40 & 30 & 90 & 0 & \textbf{100} & 40 & \textbf{80} & \textbf{100} & \textbf{62.00} \\
    Spirit-v1.5 \cite{spiritai2026spiritv15}& 0 & 0 & \textbf{50} & \textbf{80} & 90 & 0 & \textbf{100} & 20 & 60 & 70 & 51.00 \\
    GigaBrain \cite{gigabrainteam2025gigabrain0worldmodelpoweredvisionlanguageaction}& 0 & 0 & \textbf{50} & 40 & \textbf{100} & 10 & \textbf{100} & \textbf{60} & 50 & \textbf{100} & 51.67 \\
    $\pi_{0.5}$ \cite{intelligence2025pi05visionlanguageactionmodelopenworld}& 0 & 20 & 30 & \textbf{80} & 80 & \textbf{50} & \textbf{100} & 10 & 20 & \textbf{100} & 42.67 \\
    wall-oss \cite{zhai2025ignitingvlmsembodiedspace}& 0 & 0 & 10 & 70 & 70 & 20 & 70 & 10 & 10 & 20 & 35.33 \\
    $A_1$\textsuperscript{§} & 0 & 0 & 0 & 50 & 30 & 20 & 80 & 0 & 0 & 40 & 29.00 \\
    $\pi_{0}$ \cite{black2024pi_0}& 0 & 20 & 40 & 50 & 70 & 30 & \textbf{100} & 10 & 10 & 20 & 28.33 \\
    X-VLA\textsuperscript{§} \cite{zheng2025xvlasoftpromptedtransformerscalable}& 0 & 0 & 30 & 70 & 40 & 0 & 90 & 0 & 0 & 90 & 21.33 \\
    RDT-1B\textsuperscript{§} \cite{liu2025rdt1bdiffusionfoundationmodel}& 0 & 0 & 10 & 20 & 0 & 0 & 50 & 0 & 0 & 20 & 15.00 \\
    \bottomrule
  \end{tabularx}
  
  \vspace{0.5em}
\end{table*}

\begin{table}[t]
\centering
\renewcommand{\arraystretch}{0.9}
\caption{Adaptive early-exit inference with $A_1$-MLP. Average accuracy, per-episode computation cost (TFLOPs) and model inference time (second) under different exit criteria $c$ on the LIBERO benchmark. $^{\dagger}$ Full-layer training. $^{\ddagger}$ Multi-exit training.} 
\label{tab:dynamic-exit}
\begin{tabular}{l| ccccc| c|c}
\toprule
Config & Spatial & Object & Goal & Long & Avg. & TFLOPs & Inf. time \\
\midrule
no exit$^{\dagger}$ &
98.3 & 99.3 & 97.0 & 88.3 & 95.8 & 243.0 & 17.5  \\

no exit$^{\ddagger}$ &
97.4 & \textbf{100.0} & 97.4 & 91.0 & 96.5 & 243.0 & 17.5 \\
\midrule

c=1.0 &
\begin{tabular}[c]{@{}c@{}}97.4 \end{tabular} &
\begin{tabular}[c]{@{}c@{}}99.8 \end{tabular} &
\begin{tabular}[c]{@{}c@{}}\textbf{97.6} \end{tabular} &
\begin{tabular}[c]{@{}c@{}}\textbf{91.4} \end{tabular} &
\begin{tabular}[c]{@{}c@{}}\textbf{96.6} \end{tabular} & 205.0 \textcolor{green}{\footnotesize (15.6\%↓)}
 & 20.6\textcolor{red}{\footnotesize ↑} \\

c=0.7 &
\begin{tabular}[c]{@{}c@{}}\textbf{98.4} \end{tabular} &
\begin{tabular}[c]{@{}c@{}}99.8 \end{tabular} &
\begin{tabular}[c]{@{}c@{}}97.4 \end{tabular} &
\begin{tabular}[c]{@{}c@{}}89.6 \end{tabular} &
\begin{tabular}[c]{@{}c@{}} 96.3 \end{tabular} &148.1 \textcolor{green}{\footnotesize (39.1\%↓)}
 & 16.5\textcolor{green}{\footnotesize ↓} \\

c=0.4 &
\begin{tabular}[c]{@{}c@{}}95.6 \end{tabular} &
\begin{tabular}[c]{@{}c@{}}98.4 \end{tabular} &
\begin{tabular}[c]{@{}c@{}}95.0 \end{tabular} &
\begin{tabular}[c]{@{}c@{}}87.0 \end{tabular} &
\begin{tabular}[c]{@{}c@{}} 94.0 \end{tabular} &100.8 \textcolor{green}{\footnotesize (58.5\%↓)}
 & 6.8\textcolor{green}{\footnotesize ↓} \\

c=0.1 &
\begin{tabular}[c]{@{}c@{}}96.2 \end{tabular} &
\begin{tabular}[c]{@{}c@{}}98.2 \end{tabular} &
\begin{tabular}[c]{@{}c@{}}94.4 \end{tabular} &
\begin{tabular}[c]{@{}c@{}}80.4 \end{tabular} &
\begin{tabular}[c]{@{}c@{}} 92.3\end{tabular} & 57.0 \textcolor{green}{\footnotesize (76.6\%↓)}
 & 5.6\textcolor{green}{\footnotesize ↓} \\
\bottomrule
\end{tabular}

\end{table}


\subsubsection{Effectiveness of Inter-Layer Truncated Flow Matching}
Under standard inference, the VLM executes a full forward pass once, while the flow-matching action head performs $\delta$ = 10 denoising steps.
In contrast, during early-exit inference, the VLM executes forward propagation layer by layer, and at each layer the action head must also perform $\delta$ = 10 denoising steps and judge by Eq.~\ref{eq:delta} until exiting at layer $i$. While this reduces the computation of the VLM, it shifts more workload to the flow-matching action head, whose denoising iterations are time-consuming.
Consequently, when $c$ = 1.0 and $\delta$ = 10, although the computational cost is greatly reduced, the inference time still increases, as shown in Table~\ref{tab:dynamic-fm}. By introducing Inter-Layer Truncated Flow Matching, the model greatly shortens the denoising process to 2 steps and leverages warm-start initialization. Each layer’s denoising begins from the previous layer’s output rather than random noise. 
Warm-start initialization encourages earlier layer exiting, reducing the per-episode inference time from 27.5 to 10.5 seconds.
This approach significantly reduces the original inference time from 40.9 to 10.5 seconds per episode, while maintaining performance. Warm-start initialization (Eq.~(\ref{eq:anchor})) can also improve the success rate from 95.4\% to 96.4\%. Compared with $A_1$-MLP, $A_1$-FM exhibits higher action similarity across different layers. When $c$=1.0, the resulting threshold already causes the model to exit at early layers.

\begin{table}[t]
\centering
\small
\renewcommand{\arraystretch}{0.9}
\caption{Adaptive early-exit inference with $A_1$-FM. Accuracy, per-episode computation cost (TFLOPs) and model inference time (second) under different exit criteria $c$ and denoising steps $\delta$ on the LIBERO benchmark. $^*$ Layer $i+1$ denoising initialized with layer $i$ output (Eq.~(\ref{eq:anchor})). $^{\dagger}$ Full-layer training. $^{\ddagger}$ Multi-exit training.} 
\label{tab:dynamic-fm}

\begin{tabular}{l| ccccc|c|c}
\toprule
$c$, $\delta$ & Spatial & Object & Goal & Long & Avg. & TFLOPs & Inf. time \\
\midrule
no exit$^{\dagger}$, 10  & 97.2 & 99.2 & 94.6 & 79.2 & 92.6 & 231.3 & 37.9 \\
no exit$^{\ddagger}$, 10  & \textbf{97.4} & 99.2 & 96.2 & 91.2 & 96.0 & 229.8 & 37.8 \\
no exit$^{\ddagger}$, 2 & \textbf{97.4} & 98.6 & 96.8 & 89.6 & 95.6 & 226.9 & 32.2 \\
\midrule
1.0, 10  & 97.2 & \textbf{99.6} & 97.0 & 91.8 & \textbf{96.4} & 150.6\textcolor{green}{\footnotesize ↓} & 40.9 \textcolor{red}{\footnotesize (7.9\%↑)} \\ 
1.0, 2  & 94.6 & 99.0 & \textbf{98.0} & 90.0 & 95.4 & 167.9\textcolor{green}{\footnotesize ↓} & 27.5 \textcolor{green}{\footnotesize (27.4\%↓)} \\ 
1.0, 2$^*$  & 95.4 & 99.0 & 97.8 & \textbf{93.2} & \textbf{96.4} & 156.8\textcolor{green}{\footnotesize ↓} & 10.5 \textcolor{green}{\footnotesize (72.3\%↓)} \\
0.8, 2$^*$ & 96.6 & 98.6 & 94.8 & 88.2 & 94.6  & 116.8\textcolor{green}{\footnotesize ↓} & 9.0 \textcolor{green}{\footnotesize (76.3\%↓)}\\

\bottomrule
\end{tabular}

\end{table}

\subsubsection{Generalization Experiments}
When directly evaluating our $A_1$-FM (no exit$^{\ddagger}$, $\delta$=10), which was trained on the standard LIBERO dataset, on the more challenging LIBERO-Plus benchmark \cite{fei25libero-plus}. Despite significant distribution shifts in object layouts, language instructions, textures, and lighting conditions, the model achieved a robust success rate of 75.3\%.
This outperforms OpenVLA-OFT, $\pi_0$ , and $\pi_0$-FAST, demonstrating its superior zero-shot transfer capabilities. 

\begin{table}[h]
\centering
\caption{Zero-shot results on LIBERO-Plus benchmark.}
\label{tab:dynamic-fm-libero-plus}
\begin{tabular}{l| ccccc|c|c}
\toprule
Method  & Spatial & Object & Goal & Long & Avg. & TFLOPs & Inf. time (s) \\
\midrule
OpenVLA & 19.4 & 14.0 & 15.1 & 14.3 & 15.6 & - & -\\
OpenVLA-OFT & 84.0 & 66.5 & 63.0 & \textbf{66.4} & 69.6 & - & -\\
$\pi_0$ & 60.7 & 61.4 & 44.9 & 48.4 & 53.6 & - & -\\
$\pi_0$-FAST & 74.4 & 72.7 & 57.5 & 43.4 & 61.6 & - & -\\
$A_1$-FM  & \textbf{86.6} & \textbf{80.0} & \textbf{66.8} & 58.0 & \textbf{75.3} & 297.1 & 36.1 \\
\bottomrule
\end{tabular}
\label{tab:libero-plus}
\end{table}

\section{Conclusion}

In this paper, we introduced $A_{1}$, an adaptive truncated Vision-Language-Action (VLA) model. $A_1$ achieves excellent performance on various simulation environments and real robots through large-scale pre-training on open-source visual language data and robot action data. Simultaneously, it also possesses inference acceleration capabilities, effectively alleviating the challenge of VLA requiring massive computing power while maintaining performance.

\section{Acknowledgements}
This work is supported by National Key Research and Development Program of China(2024YFE0203100), Scientific Research Innovation Capability Support Project for Young Faculty (No.ZYGXQNJSKYCXNLZCXM-I28), National Natural Science Foundation of China (NSFC) under Grants No.62476293 and No.62372482,  and General Embodied AI Center of Sun Yat-sen University.

\clearpage
\newpage
\bibliographystyle{assets/plainnat}
\bibliography{main}

\clearpage
\suppressfloats[t]
\appendix
\setcounter{page}{1}
\phantomsection
\maketitlesupplementary

\centerline{\textbf{SUMMARY OF THE APPENDIX}}

This appendix contains additional details for this paper. The appendix is organized as follows:

\begin{itemize}
    \item \S\ref{limitations} provides \textbf{Limitations  and Future Work} of our work.
    
    \item \S\ref{method_details} shows more \textbf{Method Details}.
    
    \item \S\ref{more_results} provides \textbf{More Experiment Results}.

\end{itemize}

\section{Limitations and Future Work}
\label{limitations}
\begin{figure*}[t]
  \centering \includegraphics[width=0.88\textwidth]{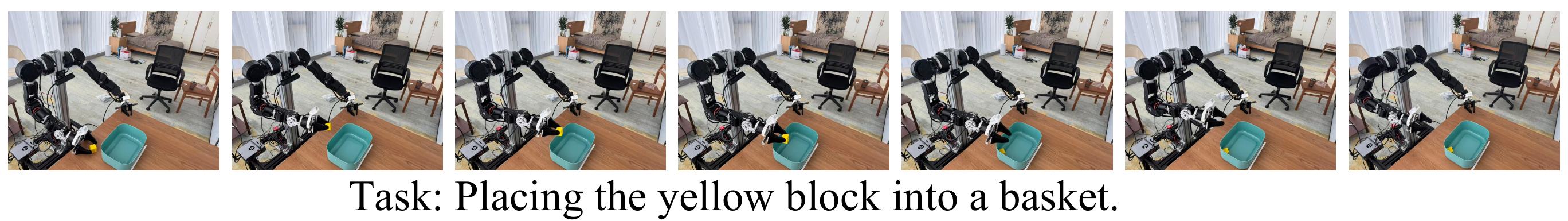} 
  \caption{The $A_1$ is deployed on our self-developed dual-arm platform OpenArm.} 
  \label{fig:wujie_arm} 
\end{figure*}

In this study, the $A_1$ model introduces affordance for pre-training, which lays the foundation for the initial performance improvement of the model. However, the current approach still faces several limitations. First, the pre-training process relies on labeled affordance datasets, which restricts the sources and scale of data. Future research could explore unsupervised learning methods to automatically mine affordance information for pre-training by leveraging the robot's data and human behavior videos. Second, the current method primarily depends on imitation learning. Although it can replicate human behavior patterns to some extent, cumulative errors gradually accumulate during processing, resulting in suboptimal operational accuracy of the model. Subsequent research could consider incorporating reinforcement learning mechanisms to dynamically adjust the model's behavioral strategies through continuous interaction and feedback with the environment, thereby enhancing the model's robustness and operational accuracy.

For adaptive early-exit inference, it is necessary to run through the training set once to compute the layer-wise action discrepancies. This is equivalent to introduce a small amount of additional training time, but it is not a major issue since the model achieves substantial acceleration during inference.

Additionally, although we have accelerated the model's inference, the synchronization of inference and execution, coupled with network latency issues between the cloud server and the local robotic arm, still leads to lag. Therefore, how to enhance the smoothness of manipulation through asynchronous execution methods is another issue that warrants further investigation.

We are also building our own dual-arm mobile control platform. The dual-arm platform offers versatile manipulation with models featuring 8 DoF in each arm, capable of precise tasks in dynamic environments. It supports master-slave arm control for high-precision teleoperation and is equipped with a mobile base, enabling it to perform mobile manipulation tasks. With a payload range of 3-5 kg, it utilises high-performance RGB cameras (IMX258) for 3D vision. $A_1$ has been successfully deployed in the platform, and the demonstration is shown in Figure \ref{fig:wujie_arm}. 

\section{Method Details}
\label{method_details}
For the probability distribution used in early-exit inference, we follow \cite{yue2024deer} and support three types of distributions.
The exponential distribution is described in detail in the main body of the paper. The Gaussian distribution emphasizes exits near a ``cente'' index 
$c$:
    \begin{equation}
            p_k \propto \exp\!\left(-\frac{(k - c)^2}{2\sigma^2}\right), 
    \quad c = \text{exit\_criterion}.
    \end{equation}
The Gamma distribution provides a skewed allocation controlled by the ``shape'' parameter:
    \begin{equation}
            p_k \propto \text{GammaPDF}(k; \alpha, \text{scale}), 
    \quad \alpha = \text{exit\_criterion}.
    \end{equation}




\section{Training Details}
\begin{table}[h]
  \centering
  \caption{Hyperparameters for pretraining $A_1$. }
  \label{tab:pretrain_hyperparams}
  \begin{tabular}{|l|l|}
    \hline
    \textbf{Configuration} & \textbf{Value} \\
    \hline
    Optimizer & AdamW \\
    Batch size & 1024 \\
    Total training steps & 200K \\
    \hline
    \multicolumn{2}{|c|}{\textbf{Learning Rates}} \\
    \hline
    ViT backbone & $0$ (frozen) \\
    VLM components & $5 \times 10^{-6}$ \\
    Action head & $5 \times 10^{-5}$ \\
    \hline
    \multicolumn{2}{|c|}{\textbf{Training Schedule}} \\
    \hline
    Warmup steps & 2,000 \\
    Freeze steps (VLM) & 1,000 \\
    LR decay & Cosine annealing \\
    \hline
    \multicolumn{2}{|c|}{\textbf{Data Augmentation}} \\
    \hline
    State mask probability & $0.5$ \\
    Visual augmentation & Random erasing, Sharpening \\
    \hline
  \end{tabular}
\end{table}

\begin{table}[h]
  \centering
  \caption{Fine-tuning hyperparameters for A1 on downstream tasks. All tasks use AdamW optimizer, ViT frozen (LR=$0$), VLM LR=$5\times 10^{-6}$, Action Head LR=$5\times 10^{-5}$, and visual augmentation (Erasing, Sharpening). The state mask probability is set to $0.5$ for all fine-tuning tasks unless otherwise specified.}
  \label{tab:finetune_hyperparams}
  \begin{tabular}{lccc}
    \toprule
    \textbf{Task / Benchmark} & \textbf{Batch Size} & \textbf{Training Steps} & \textbf{State Mask Prob} \\
    \midrule
    LIBERO & 128 & 50K & $0.0$ \\
    VLABench & 64 & 50K & $0.0$ \\
    RoboChallenge (Aloha) & 64 & 100K & $0.3$ \\
    RoboChallenge (ARX5) & 32 & 50K & $0.3$ \\
    RoboChallenge (UR5) & 64 & 50K & $0.3$ \\
    RoboChallenge (Franka) & 64 & 50K & $0.3$ \\
    \bottomrule
  \end{tabular}
\end{table}

We adopt a two-stage training pipeline consisting of large-scale pretraining followed by task-specific finetuning. 

\textbf{Pretraining.} As summarized in Table~\ref{tab:pretrain_hyperparams}, we employ the AdamW optimizer with a global batch size of 1024 for 200K total steps. The Vision Transformer (ViT) backbone remains frozen throughout (learning rate $0$), while the VLM components are initialized with a learning rate of $5\times 10^{-6}$ and the action head with $5\times 10^{-5}$. We apply a warmup strategy linearly increasing the learning rate from zero over 2,000 steps; notably, the VLM backbone is frozen (zero learning rate) during the first 1,000 steps to prevent catastrophic forgetting of pretrained capabilities, after which all trainable parameters follow cosine annealing decay. Data augmentation includes random erasing and sharpening, with a state mask probability of $0.5$ applied to proprioceptive states.

\textbf{Finetuning.} For downstream tasks, we maintain the optimizer and learning rate configuration (ViT frozen at $0$, VLM at $5\times 10^{-6}$, action head at $5\times 10^{-5}$) but adjust batch sizes and training steps according to task complexity, as detailed in Table~\ref{tab:finetune_hyperparams}. LIBERO uses batch size 128 for 50K steps with state mask probability $0.0$; VLABench uses batch size 64 for 50K steps with state mask $0.0$; RoboChallenge tasks vary---Aloha trains for 100K steps with batch size 64 and state mask $0.3$, while ARX5, UR5, and Franka train for 50K steps with batch sizes 32, 64, and 64 respectively, all using state mask probability $0.3$.

\section{More Experiment Results}
\label{more_results}

\vspace{-0.5cm}
\begin{table}[t]
\centering
\caption{Real-world experiments of adaptive early-exit inference based on AgiBot. Average accuracy and computation reduction ratio (compute↓) under different exit criterion c for model $A_1$-FM with $\delta$=2. $^{\dagger}$ Full-layer training.} 
\label{tab:exit-real}
\begin{tabular}{l lc}
\toprule
Config & Metric & Pick glue \\
\midrule
no exit$^{\dagger}$  & accuracy & 80 \\ 
\midrule

 c=1.0   &
\begin{tabular}[c]{@{}l@{}}accuracy \\ compute↓\end{tabular} & 
\begin{tabular}[c]{@{}c@{}} 70 \\ ↓49.3\end{tabular} \\

c=0.8 &
\begin{tabular}[c]{@{}l@{}}accuracy \\ compute↓\end{tabular} &
\begin{tabular}[c]{@{}c@{}} 70 \\ ↓64.7\end{tabular} \\

c=0.4 &
\begin{tabular}[c]{@{}l@{}}accuracy \\ compute↓\end{tabular} &
\begin{tabular}[c]{@{}c@{}} 80\\↓84.6 \end{tabular} \\
\bottomrule
\end{tabular}
\end{table}

\begin{figure*}[h]
  \centering \includegraphics[width=1\textwidth]{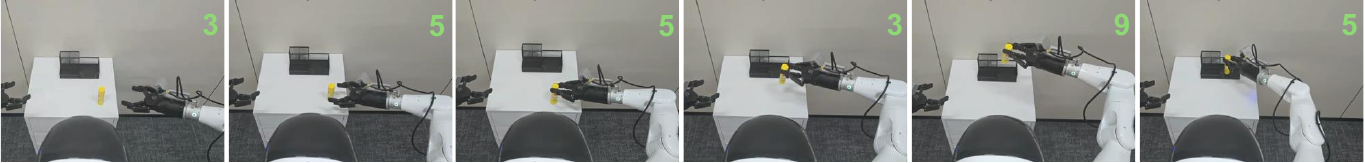} 
  \caption{AgiBot real-world example execution process of $A_1$-FM for early-exit inference with the exit criterion $c=0.4$. Green numbers indicate the layer indices where the $A_1$ model exits (the model has 28 layers in total). Task: \textit{Pick up the glue and put it into the pen holder.}} 
  \label{fig:real-exit} 
\end{figure*}
\textbf{Real-world experiments of early-termination inference}
We evaluated the adaptive early-exit inference on the AgiBot real-world task.
As shown in Table \ref{tab:exit-real}, when reducing the number of executed layers to accelerate inference, the model achieves nearly the same accuracy as full-parameter inference. When ( $c$ = 0.4 ), computation of main LLM is reduced by 84.6\% while maintaining high accuracy.
As illustrated in Figure \ref{fig:real-exit}, the model typically exits at the 3rd or 5th layer during inference.

\end{document}